# Quantizing YOLOv7: A Comprehensive Study

Mohammadamin Baghbanbashi
*School of Electrical and Computer Engineering*
*Shiraz University*
Shiraz, Iran
amin.bghb7@gmail.com

Mohsen Raji
*School of Electrical and Computer Engineering*
*Shiraz University*
Shiraz, Iran
mraji@shirazu.ac.ir

Behnam Ghavami
*Department of Computer Engineering*
*Shahid Bahonar University of Kerman*
Kerman, Iran
ghavami@uk.ac.ir

*Abstract*—YOLO is a deep neural network (DNN) model presented for robust real-time object detection following the one-stage inference approach. It outperforms other real-time object detectors in terms of speed and accuracy by a wide margin. Nevertheless, since YOLO is developed upon a DNN backbone with numerous parameters, it will cause excessive memory load, thereby deploying it on memory-constrained devices is a severe challenge in practice. To overcome this limitation, model compression techniques, such as quantizing parameters to lower-precision values, can be adopted. As the most recent version of YOLO, YOLOv7 achieves such state-of-the-art performance in speed and accuracy in the range of 5 FPS to 160 FPS that it surpasses all former versions of YOLO and other existing models in this regard. So far, the robustness of several quantization schemes has been evaluated on older versions of YOLO. These methods may not necessarily yield similar results for YOLOv7 as it utilizes a different architecture. In this paper, we conduct in-depth research on the effectiveness of a variety of quantization schemes on the pre-trained weights of the state-of-the-art YOLOv7 model. Experimental results demonstrate that using 4-bit quantization coupled with the combination of different granularities results in ~3.92x and ~3.86x memory-saving for uniform and non-uniform quantization, respectively, with only 2.5% and 1% accuracy loss compared to the full-precision baseline model.

## I. Introduction

Object detection is a computer vision technique for classifying and localizing existing objects in an image, performed with deep neural networks (DNNs). Inspired by and imitating human vision, computers can perform object detection tasks such as autonomous driving, object tracking, biometrics authentication, and pedestrian counting in crowded areas [11].

There are multiple object detection models like the R-CNN family and YOLO. R-CNN, Fast R-CNN, and Faster R-CNN are two-stage algorithms that first identify candidate regions prone to the presence of objects and then classify those areas into object classes. Despite providing high accuracy, the two-stage detection approach increases inference time and prevents real-time detection. In contrast, YOLO is a real-time object detection model, which frames the detection problem in the form of a regression problem [1][2], i.e., it detects objects in only a single stage, costing one forward propagation. YOLO simultaneously determines object bounding boxes and class probabilities in one stage [2], making it exceptionally fast. It outperforms other real-time object detectors in terms of speed and accuracy by a wide margin [2].

The latest version of YOLO, YOLOv7, is one of the most robust object detection models that achieves state-of-the-art performance in trade-offs between accuracy and speed [1]. YOLOv7 beats all previous versions of YOLO and other existing models in both speed and accuracy in the range of 5 FPS to 160 FPS [1]. However, similar to the earlier versions of YOLO, YOLOv7 is built based on deep neural networks and contain millions of parameters, making it too memory intensive. Therefore, deploying them on memory-limited gadgets and resource-constrained devices is challenging.

To tackle this issue, many research studies have been conducted so far on the compression of the YOLO model. Yang Hua et al., have proposed a dynamic balance quantization method to deploy YOLOv3 on integer-arithmetic-only hardware, which does not require fine-tuning or training data [3]. It quantizes weights to 8-bit integers (INT8) and biases to INT16 while losing 0.5% accuracy. Jun Haeng Lee et al., introduced an 8-bit linear quantization technique without relying on fine-tuning, in which the feature maps and the parameters across individual channels rather than layers get quantized to account for the inter-channel variations in the quantization range [8]. This method leads to negligible accuracy loss (less than 1%) in YOLOv2, in contrast to full-precision parameters. Since the YOLOv7 model, compared to other versions, employs a different architecture and parameter precision, it is not clear whether previous works reach the same results for YOLOv7.

In this paper, we present a comprehensive study of the performance of different quantization methods on the state-of-the-art YOLOv7 model. For this purpose, affine quantization [5][6] as a uniform quantization and Piece-wise Linear Quantization (PWLQ) [4] as a non-uniform quantization method are applied to the pre-trained weights of the YOLOv7 model. Afterward, we proceed further and investigate the effect of different granularities [5][7][9][10], such as channel-wise and filter-wise quantization approaches, on the accuracy and memory-saving ratio of the quantized model. Experimental results show that 4-bit affine quantization can lead to ~3.93x memory-saving with only about 3.4% loss in mean average precision (mAP), whereas using 4-bit PWLQ ~3.88x memory-saving can be achieved with only 1.1% accuracy loss. In addition, the mixture of different granularities results in ~3.92x and ~3.86x memory-saving for 4-bit uniform and non-uniform quantization, respectively, while incurring only 2.5% and 1% loss of accuracy. The results imply that non-uniform quantization generally surpasses the uniform scheme in terms of accuracy, albeit with a slightly more reduction in the memory-saving ratio.

Our main contributions can be summarized as follows:

- We review two main quantization strategies: uniform and non-uniform, along with their concepts, mathematical formulas, and subcategories.

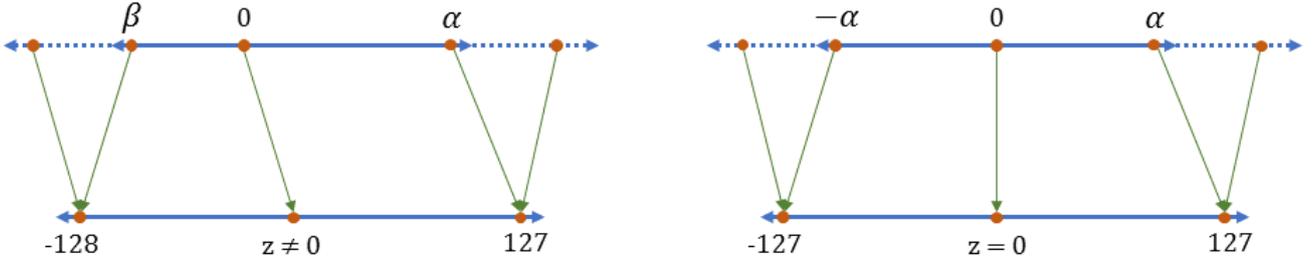

Fig. 1. Illustration of affine quantization (left) and restricted range symmetric quantization (right). Unlike the symmetric approach, affine quantization adopts an asymmetric and tighter clipping range. Thus, affine quantization is a better approach to dealing with imbalanced data.

- We classify different levels of granularity and explain them.
- We integrate each quantization method with different granularities and present an evaluation on the YOLOv7.

## II. QUANTIZATION PRINCIPLES

In this section, we first introduce two primary quantization methods in Section II.A-II.B, followed by quantization granularities in Section II.C. Later, these approaches are applied to compress YOLOv7.

### A. Uniform Quantization

Generally, in quantization, full-precision real numbers $r$ (e.g., 32-bit floating-point, FP32) are mapped into lower-precision, discrete values (e.g., INT8). In uniform quantization, the spacing of quantization levels is distributed uniformly. $[\beta, \alpha]$ denotes the permissible input range (i.e., clipping range), where inputs outside of this range are clipped to the closest bound, $\beta$ or $\alpha$ [5][6]. Two major uniform schemes can be defined due to diverse clipping ranges $[\beta, \alpha]$:

*1) Affine Quantization:* An ordinary choice for the clipping bounds is $\beta = min(r)$ and $\alpha = max(r)$. As the permissible input range is not necessarily symmetric around 0 (i.e., $-\beta \neq \alpha$), this approach is called affine or asymmetric. As illustrated in Fig. 1 (left), the affine quantization maps the real value $r \in [\beta, \alpha]$ to the quantization domain (i.e., quantized value) $\{-2^{k-1}, -2^{k-1} + 1, \ldots, 2^{k-1} - 1\}$, where $k$ is the number of quantization bits. The affine quantization function can be defined as [5][6]:

$$r_q = Q(r) = clip\left(round\left(\frac{r}{s} + z\right), -2^{k-1}, 2^{k-1} - 1\right) \quad (1)$$

$$clip(r, min, max) = \begin{cases} min, & r < min \\ r, & min \leq r \leq max \\ max, & max < r \end{cases} \quad (2)$$

where the quantization function $Q$ maps the real value $r$ to its corresponding quantized value, and the *round()* function rounds a value to its closest integer. The quantization function has two main parameters. $s$ is the scaling factor, and zero-point $z$ is the value to which the real value "0" is mapped. These parameters can be described using the following equations:

$$s = \frac{\alpha - \beta}{2^k - 1} \quad (3)$$

$$z = -round\left(\frac{\beta}{s}\right) - 2^{k-1} \quad (4)$$

The quantized value $r_q$ can be approximated to the real value $r$ through a dequantization transformation:

$$\tilde{r} = (r_q - z) \times s \quad (5)$$

Note that $\tilde{r} \neq r$, which causes quantization error.

*2) Symmetric Quantization:* In this scheme, a symmetric range is defined as the clipping range. A common option for clipping bounds is $-\beta = \alpha = max(|r|)$. Compared with symmetric quantization, affine quantization is a better choice for quantizing imbalanced inputs (e.g., non-negative data), as it mainly provides tighter clipping ranges. In this type of quantization, because of the symmetric clipping range, the zero-point $z$ is equal to 0. There are two variants of symmetric quantization. In "restricted range" symmetric quantization, the clipping range (i.e., input range) as well as quantization domain (i.e., output range) are symmetric around 0. Consequently, a single value is discarded from the output range to preserve symmetry, i.e., the usable quantization domain would be {-127, -126, …, 127} for INT8, where -128 is not used (Fig. 1, right). In addition, the scaling factor $s$ is considered to be $\alpha/2^{k-1} - 1$. As a result, for the "restricted range" the quantization function is as follows[5][6]:

$$r_q = Q(r) = clip\left(round\left(\frac{r}{s}\right), -2^{k-1} + 1, 2^{k-1} - 1\right) \quad (6)$$

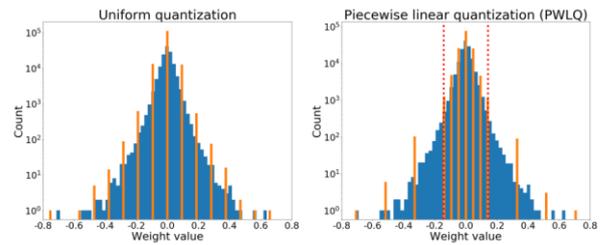

Fig. 2. Difference between uniform quantization and PWLQ. Since PWLQ quantizes the high-density central region with a higher resolution, it incurs less quantization error than the uniform approach [4].

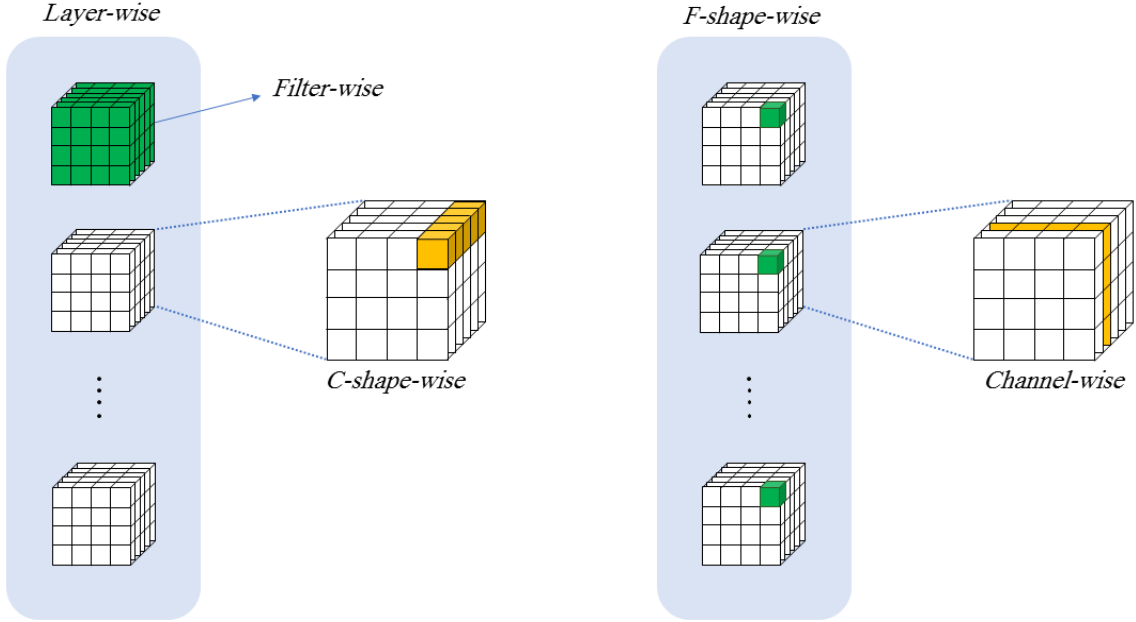

Fig. 3. Illustration of different quantization granularities for quantizing a convolutional layer. The figure shows layer-wise, filter-wise, C-shape-wise, F-shape-wise, and channel-wise granularities. Each level of granularity brings a unique level of precision due to the variance of the data.

However, in "full range" the output domain is asymmetric to use the full range, i.e., {-128, -127, …, 127} for INT8. Accordingly, the scaling factor $s$ would be $2\alpha/2^k - 1$. Therefore, we can define "full range" quantization function as [5]:

$$r_q = Q(r) = clip\left(round\left(\frac{r}{s}\right), -2^{k-1}, 2^{k-1} - 1\right) \quad (7)$$

For the both variants, dequantization operation is $\tilde{r} = r_q \times s$.

### B. Non-uniform Quantization

In non-uniform quantization, the distance between quantization levels (i.e., quantized values) is non-uniform. So, compared with uniform quantization, the non-uniform approach would be able to diminish quantization error since it divides the clipping range unequally [4][5].

*1) Piece-wise Linear Quantization (PWLQ):* Jun Fang et al., designed this non-uniform approach to benefit from the bell-shaped distribution of the weights and minimize quantization error [4]. As illustrated in Fig. 2 (right), this method divides the clipping range into two non-overlapping intervals. The intensive center region includes low-magnitude values, and the sparse tail region contains high-magnitude values. Assuming the clipping range is [-$m$, $m$], where $m = max(|r|)$, a break-point $p$ is chosen such that the clipping range is split as $Tail = [-m, -p) \cup (p, m]$ and $Center = [-p, p]$.

$2^k$ quantization levels are allocated to each of the regions. Both regions comprise a negative and a positive interval (i.e., piece), where on each of these four intervals $(k-1)$-bit uniform quantization is applied [4]. Practically, this is done by performing a $(k-1)$-bit affine quantization on each positive and negative tail and a $k$-bit full range symmetric quantization on the whole central region. So, the PWLQ function can be defined as follows:

$$PWLQ(r, k, m, p)$$
$$= \begin{cases} affine(r, k-1, \beta = -m, \alpha = -p), & r \in negative\ Tail \\ symmetric(r, k, \beta = -p, \alpha = p), & r \in Center \\ affine(r, k-1, \beta = p, \alpha = m), & r \in positive\ Tail \end{cases} \quad (8)$$

The value of break-point $p$ plays a vital role in the quantization performance. Jun Fang et al., have formulated the optimal break-point $p^*$ in the form of an optimization problem, which is not easy to solve [4]. However, they have proposed a fast approximation of $p^*/m = ln(0.8614m + 0.6079)$ that estimates $p^*$ fairly close to the optimal value.

### C. Quantization Granularity

One of the pivotal stages in quantization that has a decisive impact on its performance is selecting a granularity. Quantization granularity determines in which groups the weights have to be quantized; i.e., it determines which weights share the same quantization parameters. In CNN-based models, such as YOLOv7, the activations of a layer would be convolved with lots of various convolutional filters containing hundreds of parameters (i.e., weights). Since there is a vast number of weights in a layer, it would be essential to categorize the weights as different groups to perform quantization across each group separately. Thus, the weights belonging to the same group will have the same quantization parameters (e.g., scaling factor, zero-point). Therefore, quantization granularity is a determining factor in both accuracy and memory-saving ratio. Fig. 3 shows a number of granularity schemes that can be specified as follows [5][7][9][10]:

*1) Layer-wise Quantization:* In this method, the same clipping range $[\beta, \alpha]$ is applied to the entire layer. Hence, all

TABLE I. COMPARISON OF UNIFORM AND NON-UNIFORM QUANTIZATION WITH SINGLE-GRANULARITY APPROACH

| Quantization Method | Granularity | Bits | Precision | Recall | mAP | Memory-Saving |
|---|---|---|---|---|---|---|
| Yolov7 | | 16 | 0.724 | 0.635 | 0.691 | |
| Uniform | F-shape-wise | 4 | 0.704 | 0.579 | 0.635 | ~ 3.93 |
| Non-Uniform | F-shape-wise | 4 | 0.733 | 0.592 | 0.659 | ~ 3.87 |
| Uniform | Filter-wise | 4 | 0.605 | 0.464 | 0.491 | ~ 3.97 |
| Non-Uniform | Filter-wise | 4 | 0.723 | 0.615 | 0.672 | ~ 3.94 |
| Uniform | C-shape-wise | 4 | 0.708 | 0.604 | 0.657 | ~ 3.93 |
| Non-Uniform | C-shape-wise | 4 | 0.715 | 0.632 | 0.680 | ~ 3.88 |
| Uniform | Channel-wise | 4 | 0.686 | 0.573 | 0.621 | ~ 1.68 |
| Non-Uniform | Channel-wise | 4 | 0.701 | 0.632 | 0.672 | ~ 1.68 |

TABLE II. COMPARISON OF UNIFORM AND NON-UNIFORM QUANTIZATION WITH MULTI-GRANULARITY APPROACH

| Quantization Method | Granularity | Bits | Precision | Recall | mAP | Memory-Saving |
|---|---|---|---|---|---|---|
| Yolov7 | | 16 | 0.724 | 0.635 | 0.691 | |
| Uniform | Mix of 4 options | 4 | 0.732 | 0.610 | 0.675 | ~ 2.91 |
| Non-Uniform | Mix of 4 options | 4 | 0.714 | 0.636 | 0.682 | ~ 2.89 |
| Uniform | Mix of 3 options | 4 | 0.713 | 0.610 | 0.666 | ~ 3.92 |
| Non-Uniform | Mix of 3 options | 4 | 0.712 | 0.634 | 0.681 | ~ 3.86 |

the weights among a layer share the same quantization parameters. However, it generally results in poor accuracy since the clipping range of an individual filter can broadly differ from another [5][7].

*2) Filter-wise Quantization:* The widely used filter-wise quantization considers all the weights of a convolutional filter as a group and quantizes each filter individually [9]. As a consequence, in most cases, it leads to a better performance than layer-wise quantization [5].

*3) Channel-wise Quantization:* This approach quantizes each channel of a convolutional filter independently, i.e., exclusive quantization parameters are assigned to each channel. This method provides a satisfactory quantization accuracy for YOLOv7. However, the main downside is that many convolutional layers in YOLOv7 cannot get quantized using this granularity since they have a kernel size of (1, 1), which means there exists only a single value in each channel of a filter (e.g., conv5 and conv44 layers). As a result, channel-wise quantization for YOLOv7 hurts the memory-saving ratio.

*4) Shape-wise Quantization over Filters (F-shape-wise):* A convolutional layer usually contains tens of convolutional filters. By applying this granularity, weights within different filters of the same layer having the same height, width, and channel number get quantized with the same quantization parameters [10]. This is analogous to generating a filter for scaling factor and zero-point in each layer, each element of which corresponds to exactly one weight per filter.

*5) Shape-wise Quantization over Channels (C-shape-wise):* In general, a convolutional filter consists of multiple channels. The C-shape-wise approach quantizes the weights placed in the same position across different filter channels as a group [9]. The importance of this approach can be seen in the fact that it outperforms other quantization granularities for YOLOv7, which will be shown in Section III.B.

As illustrated in Fig. 3, each granularity deals with a distinct distribution of data, resulting in a different precision and memory-saving ratio in the quantized model.

## III. EXPERIMENTAL RESULTS

In this section, several experiments are conducted to study the effects of different quantization methods on the YOLOv7 model. The results are evaluated on the MS COCO (val2017) dataset.

### A. Experiment Setup

We perform 4-bit uniform and non-uniform quantization with four different quantization granularities filter-wise, channel-wise, F-shape-wise, and C-shape-wise. As discussed in Section II, we adopt affine quantization as a uniform and PWLQ as a non-uniform quantization method in our experiments. In any of the experiments, we don't quantize batch normalization modules as they are prone to high quantization errors because of their uniform-like distribution. All experiments are performed in the Pytorch 1.12.1 framework on a remote server containing an RTX6000 GPU, 64 GB RAM, and a 12-core CPU.

### B. Results on YOLOv7

Table I compares different quantization methods with various granularities in terms of precision, recall, mAP, and memory-saving ratio. In the YOLOv7 model, the weight parameters are stored in FP16 format. After applying uniform quantization (i.e., affine quantization) with 4-bit integers on the baseline model, ~3.93x memory-saving can be obtained with only about 3.4% loss in mAP. 4-bit non-uniform quantization (i.e., PWLQ) can also bring up to ~3.88x memory-saving with only 1.1% accuracy loss. As the results indicate, since the pre-trained weights of the YOLOv7 model almost follow Gaussian or Laplacian bell-shaped distribution with mean ≈ 0, PWLQ provides better accuracy in contrast to affine quantization.

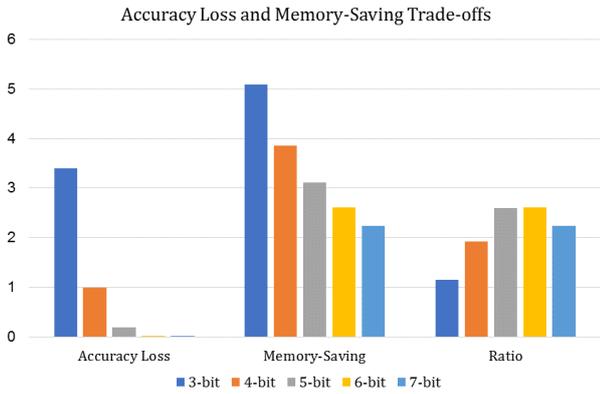

Fig. 4. Comparison of *memory saving*/(*accuracy loss* + 1) ratio over several *k*-bit PWLQ with combination of filter-wise, F-shape-wise, and C-shape-wise granularities.

### C. Further Exploration

In this section, the trade-off between memory-saving and accuracy loss is further discussed by combining diverse quantization granularities and choosing a various number of quantization bits.

*1) Combination of Various Granularities:* We can pick the most appropriate granularity option for each module by applying all quantization granularities to its weights and comparing the resulting quantization errors. Then we can further reduce the accuracy loss of 4-bit affine quantization and PWLQ to 1.6% and 0.9%, respectively. Meanwhile, the overall memory-saving ratio decreases due to different granularity options assigned to each module. Table II compares uniform and non-uniform quantization from a multi-granularity viewpoint. In the mixture of 3 granularity options, the channel-wise approach is not considered since it hurts the memory-saving ratio.

*2) K-bit Quantization:* Memory-saving ratio and accuracy are inversely related, and the higher one is, the lower the other is likely to be. We present the ratio of *memory saving*/(*accuracy loss* + 1), as a performance metric and check that ratio for multiple *k*-bit PWLQ with a combination of filter-wise, C-shape-wise, and F-shape-wise. The reason why we added accuracy loss with 1 is that practically from 6-bit onwards, we will no longer have a loss in accuracy, i.e., accuracy loss is 0%. Fig. 4 demonstrates that the ratio reaches its peak in 5-bit and 6-bit quantization and then goes on a downward trend.

## IV. CONCLUSION

This paper presents a comprehensive study on the efficacy of different quantization methods to address the high memory usage problem in the state-of-the-art YOLOv7 model. We apply affine quantization and PWLQ as representatives of uniform and non-uniform methods, respectively, along with a variety of granularities to the pre-trained weights of YOLOv7. Empirically, C-shape-wise granularity yields the best results for both affine quantization and PWLQ among the other granularity perspectives. Using 4-bit C-shape-wise affine quantization gives rise to ~3.93x memory-saving with 3.4% mAP loss, while 4-bit C-shape-wise PWLQ achieves ~3.88x memory-saving with only 1.1% mAP loss. However, as the experimental results indicate, for both affine quantization and PWLQ, the combination of filter-wise, F-shape-wise, and C-shape-wise approaches offers even more promising performance when it comes to balancing the memory-saving ratio against accuracy loss in contrast with single-granularity viewpoints. In conclusion, it is worth mentioning that affine quantization and PWLQ coupled with the mixture of filter-wise, F-shape-wise, and C-shape-wise schemes are the most suitable choices for deploying YOLOv7, correspondingly, on mobile CPUs (or GPUs) and FPGAs.